\pdfoutput=1

\documentclass[11pt]{article}

\usepackage[final]{acl}

\usepackage{times}
\usepackage{latexsym}
\usepackage[normalem]{ulem}

\usepackage[T1]{fontenc}

\usepackage[utf8]{inputenc}

\usepackage{microtype}

\usepackage{inconsolata}

\usepackage{graphicx}

\usepackage{amsmath}
\usepackage{amssymb}
\usepackage{bm}
\usepackage{algorithm}
\usepackage{algorithmic}
\usepackage{hyperref}

\usepackage{booktabs}
\usepackage{multirow}

\title{Revealing COVID-19's Social Dynamics: Diachronic Semantic Analysis of Vaccine and Symptom Discourse on Twitter}

\author{%
  \normalsize Zeqiang Wang$^{1}$ \quad
  Jiageng Wu$^{2}$ \quad
  Yuqi Wang$^{3}$ \quad
  \textbf{Wei Wang}$^{3}$ \quad
  \\
  \normalsize \textbf{Jie Yang}$^{4}$ \quad
  \textbf{Jon Johnson}$^{5}$ \quad
  \textbf{Nishanth Sastry}$^{1}$ \quad
  \textbf{Suparna De}$^{1}$\thanks{Corresponding Author.}\\
  \small $^{1}$ University of Surrey \small $^{2}$ Zhejiang University
  \small $^{3}$ Xi'an Jiaotong Liverpool University\\
  \small $^{4}$ Harvard University $^{5}$ University College London\\
  \small \texttt{\{zeqiang.wang, s.de, n.sastry\}@surrey.ac.uk, jiagengwu@zju.edu.cn, jieynlp@gmail.com,}  \\ \small \texttt{yuqi.wang17@student.xjtlu.edu.cn, wei.wang03@xjtlu.edu.cn,} \small
  \texttt{jon.johnson@ucl.ac.uk}
}

\begin{document}
\maketitle
\begin{abstract}
Social media is recognized as an important source for deriving insights into public opinion dynamics and social impacts due to the vast textual data generated daily and the ‘unconstrained’ behavior of people interacting on these platforms. However, such analyses prove challenging due to the semantic shift phenomenon, where word meanings evolve over time. This paper proposes an unsupervised dynamic word embedding method to capture longitudinal semantic shifts in social media data without predefined anchor words. The method leverages word co-occurrence statistics and dynamic updating to adapt embeddings over time, addressing the challenges of data sparseness, imbalanced distributions, and synergistic semantic effects. Evaluated on a large COVID-19 Twitter dataset, the method reveals semantic evolution patterns of vaccine- and symptom-related entities across different pandemic stages, and their potential correlations with real-world statistics. Our key contributions include the dynamic embedding technique, empirical analysis of COVID-19 semantic shifts, and discussions on enhancing semantic shift modeling for computational social science research. This study enables capturing longitudinal semantic dynamics on social media to understand public discourse and collective phenomena.
\end{abstract}

\section{Introduction}

Social media has become an essential platform for information dissemination and public opinion expression, with over 4.5 billion global users generating vast amounts of textual data daily \cite{li2022tracking, wu2023trend, De2023}. Large-scale, wide-coverage, and real-time social media data provide a new source for computational social science and cultural analysis research, enabling insights into public opinion dynamics, social event impacts, and collective behavior patterns through large-scale text analysis \cite{stream-jamia-2024}.
However, social media text analysis faces a unique challenge: the semantic shift phenomenon. Word semantics can change to a certain extent over time due to evolving social events, user interests, cultural trends, and other factors \cite{harrigian2022problem, wang2023fusing}. 
For instance, during the COVID-19 pandemic, 
``\textit{Moderna}'' initially referred to a biotechnology company but later became associated with its COVID-19 vaccine. Failing to account for such dynamic semantic evolution can affect the accuracy and interpretability of text analysis, especially in long-term longitudinal analyses spanning months or even years.

Existing mainstream semantic shift analysis methods usually require pre-defining a set of anchor words as a reference \cite{ishihara2022semantic, montanelli2023survey}. However, these methods rely excessively on human experts' prior experiences that are usually highly subjective~\cite{sharifian2022}, and cannot comprehensively detect semantic shifts in longitudinal corpora. Moreover, social media text presents several challenges for existing semantic shift analysis methods:
\begin{enumerate}
    \item \textbf{Data Sparseness}: Social media text is highly sparse, with only a few hundred core words and categories exhibiting dynamic changes, making it difficult for statistical language models to learn robust semantic representations.
    \item \textbf{Imbalanced Distribution}: Core words have a highly imbalanced distribution. This long-tail distribution poses challenges for semantic shift modeling, requiring targeted handling of low-frequency and emerging words.
    \item \textbf{Synergistic Effects}: Topic evolution on social media often manifests as the joint shift of multiple related words. Focusing solely on isolated word changes may obscure overall topic-level trends.

\end{enumerate}

To address these challenges, this paper proposes a novel framework aimed at better capturing longitudinal semantic evolution patterns in the context of social media and providing powerful semantic analysis tools for computational social science research. The proposed end-to-end unsupervised dynamic word embedding method is based on word co-occurrence statistics and combines a dynamic update strategy to adaptively capture the trajectory of word semantic evolution over time, without the need for manually defined anchor words.
To verify the effectiveness of the method in practical applications, we applied it to a large-scale COVID-19 Twitter dataset spanning more than two years, from February 2020 to April 2022. We focused on the semantic shift phenomenon of vaccine- and symptom-related entity words and observed their semantic shift patterns at different stages of the pandemic, as well as their potential correlations with real epidemic trends.

The main contributions are summarized as follows:
1) We propose a dynamic word embedding method that can adaptively detect semantic shift phenomena in longitudinal corpora without manual annotation;
2) We integrate it with Named Entity Recognition (NER) and Large Language Model (LLM)-based entity normalization to systematically investigate the semantic shift in extensive and longitudinal social media corpora during the COVID-19 pandemic;
3) Our experimental results reveal the semantic evolution patterns of vaccine- and symptom-related topics during the COVID-19 pandemic, offering a new perspective on the study of epidemic evolution and its social impact. 
This provides a novel paradigm for leveraging computational linguistics in social and medical research.


\begin{figure*}[t!]
    \includegraphics[width=\textwidth]{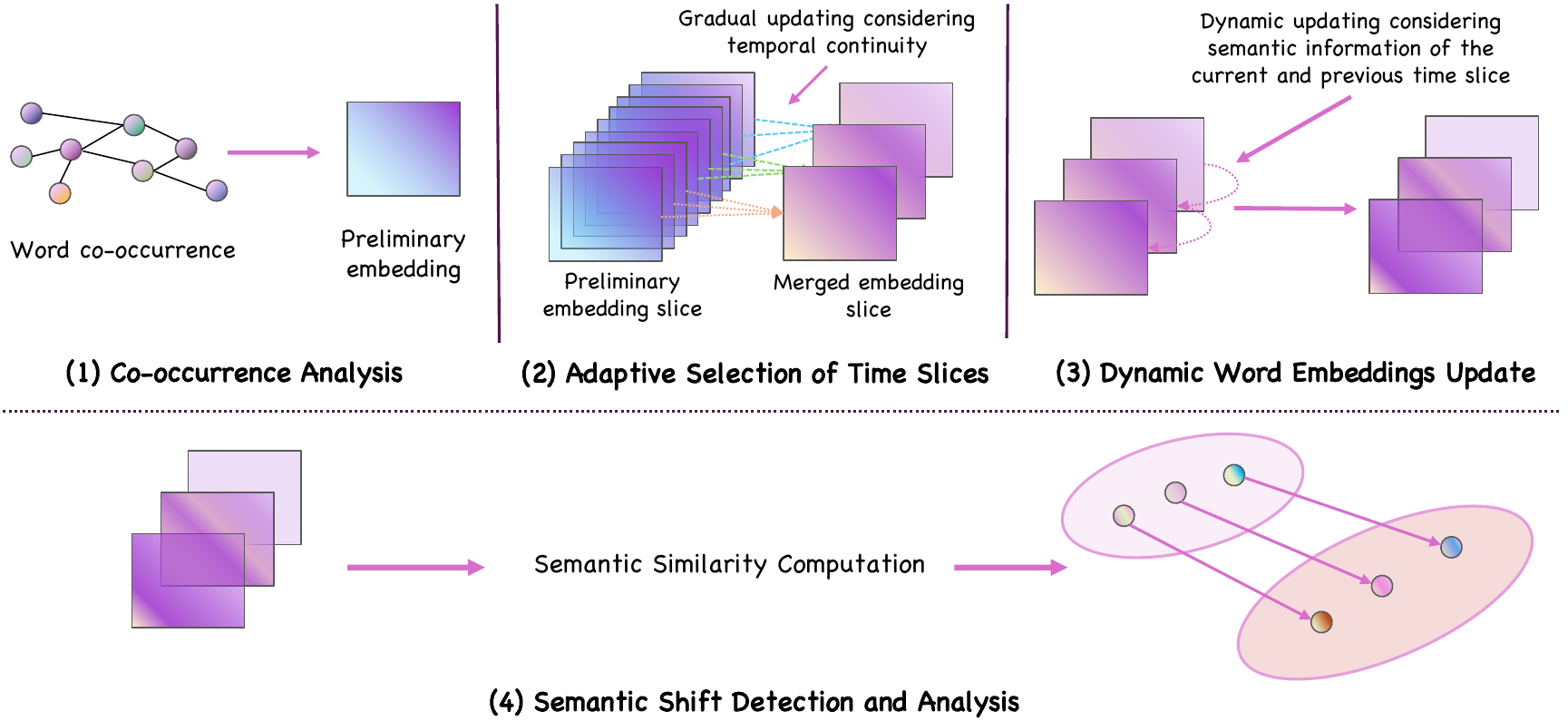}
\caption{Overall framework of the proposed unsupervised dynamic word embedding method. (1) Co-occurrence Analysis: Word co-occurrence matrices from a diachronic corpus are computed and normalized;
(2) Adaptive Selection of Time Slices: Adjacent time slices with high word co-occurrence similarity are merged adaptively;
(3) Dynamic Word Embeddings Update: Word embeddings are dynamically updated based on the current and previous time slices.
(4) Semantic Shift Detection and Analysis: Word semantic shifts are detected by embedding similarity and tracking the associations of word pairs that change over time.}
\label{fig:main}
\end{figure*}

\section{Related Work}

Semantic shift analysis is a crucial research problem in the fields of computational linguistics and natural language processing. Over time, the semantics of words evolve, reflecting changes in language usage and socio-cultural dynamics. Researchers have recently proposed various data-driven methods to automatically detect semantic shift phenomena at the word level.

Early studies on semantic shift primarily relied on word frequency statistics, analyzing the temporal variation of word frequencies in books~\cite{michel2011quantitative} and in web search queries \cite{choi2012predicting}. However, it has been noted that word frequency changes are not always directly correlated with semantic shifts \cite{kulkarni2015statistically}, with frequency information alone being insufficient to accurately capture subtle semantic variations.

Distributional semantics has been proven to be an effective tool for semantic shift analysis \cite{turney2010frequency, baroni2014don}. By leveraging word co-occurrence statistics, distributional representations of words can be learned, directly encoding their semantics as vectors. \citet{jurgens2009event} pioneered quantitative analysis of semantic shifts with word vector updates to dynamically model semantic evolution, with subsequent studies employing various word embedding algorithms, such as Latent Semantic Analysis by \citet{sagi2011tracing} and Local Mutual Information by \citet{gulordava2011distributional}.
\citet{kim2014temporal} introduced predictive word embedding models (e.g., word2vec) to the semantic shift detection task, followed by works comparing performance of different types of word embedding models \cite{hamilton2016diachronic} which found that predictive models like skip-gram with negative sampling (SGNS) outperform explicit counting methods.

Other studies~\cite{mitra2015automatic} have looked at characterizing semantic shift types, such as the emergence of novel senses, splitting and merging of word senses, which falls under fine-grained semantic shift analysis. \citet{hamilton2016cultural} explored the distinction between cultural and linguistic shifts as the two modes of semantic evolution, with the former closely related to socio-cultural changes and manifesting as global displacement in the word vector space, and the latter reflecting the internal evolution of language and characterized by changes in local neighborhoods.
Semantic shift analysis through the topic evolution lens is explored through dynamic topic models to track the temporal variation of topic distributions~\cite{wang2006topics} or temporal evolution of Google search topics~\cite{wijaya2011understanding}.


The emergence of contextualized word embeddings capable of capturing diversity and subtle changes in word semantics in pre-trained language models such as ELMo \cite{peters-etal-2018-deep} and BERT \cite{devlin2018bert} has led to the use of BERT for semantic shift detection, with a novel representation splitting method to improve detection accuracy \cite{hu2019diachronic}. Subsequent studies, such as those by~\citet{giulianelli2020analysing} have confirmed their superior performance over static ones with systematic evaluations of different types of contextualized word embeddings on this task.



Moreover, some studies focused on the interpretability of semantic shift detection. \citet{rosin2022time} proposed an interpretable semantic shift model based on attention mechanisms, explaining the reasons for word semantic evolution by visualizing attention weights. \citet{you2021unsupervised} employed causal inference methods to investigate the socio-cultural driving factors behind lexical semantic shifts. These works contribute to a deeper understanding of semantic shift phenomena from both computational and humanistic perspectives.
In comparison to the above-mentioned methods, suppressing high-frequency core words and dynamically updating weights ensure that we can detect stable semantic shifts in embeddings.

Regarding the analysis of COVID-19 discussions on social media, several studies have explored this topic from different perspectives. \citet{choudrie2021machine} investigated the role of machine learning in identifying misinformation during the pandemic, and explored how older adults interacted with online information, often finding it difficult to verify and preferring traditional media sources. \citet{wicke2020framing} focused on the metaphorical language used in social media, particularly war-related terms, showing how certain metaphors shaped public conversations around the virus and its effects. \citet{gencoglu2020causal} analyzed the relationship between pandemic trends, such as infection rates, and shifts in public sentiment on Twitter, using causal modeling to distinguish between events that influenced public attention and those that were merely correlated. These studies collectively offer insights into the dynamic nature of COVID-19 discourse online, shaped by both statistical methods and human perception.
\section{Methodology}
This section introduces the proposed unsupervised dynamic word embedding method for detecting and analyzing semantic shifts of words in detail. The overall framework is shown in Figure \ref{fig:main}. Our method leverages word co-occurrence statistics and adopts adaptive selection and time slice dynamic updating strategies to analyze and capture the evolution of word semantics over time.

\subsection{Co-occurrence Analysis}
Let $\mathcal{D} = \{\mathcal{D}^1, \mathcal{D}^2, \dots, \mathcal{D}^T\}$ denote the diachronic text corpus, where $\mathcal{D}^t$ represents the text collection of the $t$-th time slice. We collect and pre-process large-scale text data from a number of time periods, including news articles, books, social media posts, and more, to ensure the data covers a broad temporal span and is representative of each period.
We compute the word co-occurrence frequency matrix $X^t \in \mathbb{R}^{|V| \times |V|}$ on each time slice $\mathcal{D}^t$, where $|V|$ represents the size of the vocabulary. $X^t_{ij}$ represents the co-occurrence frequency of words $w_i$ and $w_j$ in the $t$-th time slice, reflecting the contextual information of words in different time periods.

To address the data sparseness, we apply a smoothing method to process co-occurrence matrices. Specifically, add-one smoothing is employed to add a small constant $\alpha$ to all elements of the matrix:
\begin{equation}
\tilde{X}^t_{ij} = X^t_{ij} + \alpha
\end{equation}
where $\tilde{X}^t$ denotes the smoothed co-occurrence matrix. This smoothing technique helps mitigate the impact of sparse data and improves the robustness of subsequent computation.
Furthermore, to handle the unbalanced distribution of core words, we apply frequency-based normalization to the co-occurrence matrices. Specifically, we use logarithmic scaling to balance the importance of core words based on their frequency:
\begin{equation}
\hat{X}^t_{ij} = \log(1 + \tilde{X}^t_{ij})
\end{equation}
where $\hat{X}^t$ denotes the normalized co-occurrence matrix. This normalization technique helps reduce the impact of extremely frequent words and gives more weight to less frequent but potentially informative words.
We pre-train word vectors using traditional static word embedding methods (such as word2vec \cite{mikolov2013efficient} or GloVe \cite{pennington-etal-2014-glove}) to obtain preliminary embedding for each time slice $t$, i.e. $W^t \in \mathbb{R}^{|V| \times d}$, where $d$ is the embedding dimension.

\subsection{Adaptive Selection of Time Slices}
We adaptively adjust the span of time slices based on the temporal granularity of word co-occurrence statistics. First, we compute a similarity measure between word co-occurrence matrices of adjacent time slices. Then, by setting a similarity threshold $\tau$, we merge adjacent time slices with a similarity score higher than the threshold, forming slices with larger time spans. This ensures the continuity of semantic evolution while reducing the number of time slices and improving computational efficiency.

We use cosine similarity between the normalized word co-occurrence matrices  $\hat{X}^t$ and $\hat{X}^{t+1}$ of adjacent time slices $t$ and $t+1$, which is given by:
\begin{equation}
\text{sim}(\hat{X}^t, \hat{X}^{t+1}) = \frac{\text{vec}(\hat{X}^t) \cdot \text{vec}(\hat{X}^{t+1})}{|\text{vec}(\hat{X}^t)|_2 |\text{vec}(\hat{X}^{t+1})|_2}
\end{equation}

where $\text{vec}(\cdot)$ denotes flattening the matrix into a vector. If the semantic similarity is greater than a predefined threshold $\tau$, the two time slices are merged into one. The corresponding word embeddings are then re-computed based on the merged corpus $\mathcal{D}^t$ and $\mathcal{D}^{t+1}$. These steps are repeated until there are no similar co-occurrence matrices for any two consecutive time slices.

\subsection{Dynamic Word Embeddings Update}
We adopt a progressively dynamic update strategy. Given the smoothed word co-occurrence matrix $\tilde{X}^t$ of the current time slice $t$ and the word embedding matrix $W^{t-1}$ learned from the previous time slice, we update the word embeddings by minimizing the following objective:
\begin{equation}
W^t = \arg\min_W \left| \hat{X}^t - WW^T \right|_F^2 + \lambda \left| W - W^{t-1} \right|_F^2
\end{equation}
where $|\cdot|_F$ denotes the Frobenius norm of a matrix, and $\lambda$ is a weight coefficient balancing the reconstruction error of the current time slice and the similarity with the embeddings from the previous time slice. 
This enables word embeddings to adapt to current semantic information while retaining previously learned semantics.

We introduce a forgetting factor $\gamma \in (0, 1]$ to control the speed at which word embeddings forget outdated semantics. Before each update, we decay the word embeddings from the previous time slice:
\begin{equation}
\tilde{W}^{t-1} = \gamma W^{t-1}
\end{equation}
Then, we substitute the decayed word embeddings $\tilde{W}^{t-1}$ into the optimization objective for updating. A smaller $\gamma$ value promotes faster forgetting of outdated semantics, while a larger $\gamma$ value helps maintain the continuity of semantics.
To maintain the global semantic stability of word embeddings, we introduce a regularization term in the dynamic update optimization objective: 

\begin{equation}
\begin{split}
W^t = &\arg\min_W \left| \hat{X}^t - WW^T \right|_F^2 \\
&+ \lambda_1 \left| W - \tilde{W}^{t-1} \right|_F^2 + \lambda_2 \left| WW^T - \bar{X} \right|_F^2
\end{split}
\end{equation}
where $\bar{X} = \frac{1}{T}\sum_{t=1}^T \hat{X}^t$, represents the average normalized word co-occurrence matrix across all time slices, and $\lambda_2$ controls the weight of global stability. By minimizing the difference between the inner product of word embeddings $WW^T$ and the average word co-occurrence matrix $\bar{X}$, this regularization term encourages word embeddings to maintain global semantic consistency across different time and suppresses excessive semantic shift.

\subsection{Semantic Shift Detection and Analysis}

To analyze semantic shift under complex interactions and capture synergies between words, we introduce higher-order co-occurrence statistics. Specifically, we calculate the cosine similarity between word pairs across different time slices.

By tracking changes in cosine similarity of word pairs across different time slices, we can uncover dynamic patterns of semantic association evolution, such as strengthening, weakening, or shifting of associations. For instance, if the cosine similarity between ``\textit{COVID-19}'' and ``\textit{vaccine}'' gradually increases over time, it suggests that their semantic association becomes stronger with the progress of vaccine development and deployment.

\section{COVID-19 Case Study}
\subsection{Dataset construction}
We leveraged large-scale Twitter data to support an in-depth analysis of the characteristics and evolution patterns of COVID-19 symptoms. We selected 471,553,966 non-retweet English tweets posted between February 1, 2020, and April 30, 2022, as the raw dataset. This time span covered multiple key stages of the pandemic, including the initial outbreak, global spread, vaccine development, and rollout. The substantial data volume and extended time range provide a comprehensive and longitudinal perspective on the pandemic's full picture.

To accurately identify self-reported symptoms mentioned in the massive tweet corpus, we designed a rigorous data preprocessing pipeline consisting of three steps:

\begin{enumerate}
    \item \textbf{Strict COVID-19 relevance filtering}: We matched each tweet to a COVID-19 vocabulary consisting of 92 core keywords, derived from an infoveillance study on COVID-19 symptoms \cite{wu2023trend}, to ensure that the filtered tweets were highly relevant to the pandemic. The vocabulary covers multiple semantic categories closely related to the pandemic, such as disease names, virus strains, symptoms, and policies.
    \item \textbf{Text cleaning and normalization}: To remove noisy and irrelevant information and improve text quality, we performed a series of text cleaning operations, including: a) removing metadata such as URLs, @usernames, and \#hashtags; b) converting tweets to lowercase; c) removing stop words, punctuation, and special characters; d) performing lemmatization to merge word variants.
    \item \textbf{Self-reported symptom matching}: To precisely locate self-reported symptom expressions in the cleaned tweets, we carefully constructed a high-coverage concept hierarchy for COVID-19 symptoms, which contains 10 body systems, 257 core symptoms, and 1,808 synonyms. We adopted a combination of exact matching and fuzzy matching methods to match each tweet against the symptom hierarchy. Exact matching could identify explicitly mentioned standard symptom terms in tweets, while fuzzy matching significantly improved the recognition coverage of colloquial and non-standard symptom expressions through synonym expansion.
\end{enumerate}

Through the three-step preprocessing pipeline, we filtered 948,478 COVID-19-related tweets containing self-reported symptoms from the original over 400 million tweets, accounting for 0.2\% of the total tweet volume. Despite the low proportion, these tweets were of rich information and high quality, providing a valuable data foundation for conducting large-scale, real-world COVID-19 symptom analysis.

\subsection{Symptom and Vaccine Entity Recognition}
To further improve the recognition of symptom and vaccine entities, we used deep learning-based named entity recognition (NER) techniques. We adopted the YATO framework \cite{wang-etal-2023-yato}, using the pre-trained covid-twitter-bert-v2 model \cite{muller2023covid}, and fine-tuned it on the METS-CoV dataset \cite{zhou2022mets} for the fine-grained NER task. The model can automatically identify two types of entities relevant to this study from tweets, including symptoms and vaccines.

We applied the fine-tuned NER model for comprehensive inference on the previously screened, approximately one million tweets on self-reported symptoms and extracted the symptom and vaccine entities. Compared to traditional rule-based or dictionary-matching methods, deep learning-based NER can better handle the colloquial, non-standard, and complex nature of social media text, thereby significantly improving the precision and recall of entity recognition. 
As shown in Table \ref{ner_result}, it achieved a high F1-score of 0.89 on METS-CoV.

\begin{table}[htbp]
\centering
\small
\begin{tabular}{cccc}
\toprule
\textbf{Entity}       & \textbf{Precision} & \textbf{Recall} & F1-\textbf{score} \\
\midrule
Symptom         & 0.86      & 0.86   & 0.87     \\
Vaccine & 0.88      & 0.93   & 0.90     \\ 
\midrule
Overall (macro)        & 0.87      & 0.90   & 0.89     \\
\bottomrule
\end{tabular}
\caption{NER Performance on METS-CoV.}
\label{ner_result}
\end{table}

\begin{table*}[]
\footnotesize
\centering
\resizebox{\textwidth}{!}{%
\begin{tabular}{clcl}
    \toprule
    \textbf{Type}                  & \textbf{Normalized Entity (ICD-11)} & \textbf{Count (freq\%)} & \textbf{Raw Entity in Social Media}                                      \\
    \midrule
    \multirow{5}{*}{\textbf{Physical}}      & Nausea or Vomitting                 & 69670 (6.80\%)          & nausea head, upset stomach, vomit, covid stomach, vom, nause,…           \\
     & Fever                 & 67108 (6.55\%) & fever, feverish, cough fever, high temp, heat …                                  \\
     & Cough                 & 55326 (5.40\%) & choking cough, sore throat coughing, chesty cough pain, bad cough, … \\
                                   & Disturbances of Smell and taste     & 53891 (5.26\%)          & lost tast smell, tasteless, couldnt smell, loss taste, loss smell, …     \\
                                   & Dizziness and giddness              & 49178 (4.80\%)          & dizzy head, dizzyer, brain feels like gone, lightheaded,feeling giddy, … \\
    \midrule
    \multirow{5}{*}{\textbf{Psychological}} & Anxiety                             & 2,121 (0.20\%)           & freaking out, nervous wreck, on edge, jittery, stressing out,…           \\
     & Depressed mood        & 2,013 (0.19\%)  & feeling down, bummed out, low spirits, in a funk, blue, …                        \\
     & Irritability          & 1,822 (0.17\%)  & cranky, snappy, grouchy, on a short fuse, annoyed, …                             \\
     & Psychomotor agitation & 1,672 (0.16\%)  & can't sit still, fidgety, restless, hyper, pacing, …                             \\
     & Aggressive behavior   & 1,423 (0.13\%)  & lashing out, losing it, going off, snapping, getting in fights, …   
                  \\
    \bottomrule
    \end{tabular}
}
\caption{Example of the correspondence between social media symptom entities based on large language model normalization and standardized symptom entities.}
\label{nn_result}
\end{table*}

Through entity recognition on the full set of tweets, we extracted a total number of 93,418 symptom entities and 7,313 vaccine entities. It is worth noting that the extracted entities exhibited diverse forms, including standard symptom terms such as ``\textit{fever}'' and ``\textit{shortness of breath}''; non-standard descriptions such as ``\textit{feeling feverish}'' and ``\textit{out of breath}''; formal vaccine names like ``\textit{COVID19 Vaccine}''; and colloquial expressions like ``\textit{pfizer vaccine}''  and ``\textit{az jab}''. This highlights the flexibility and diversity of semantic expressions in social media text and confirms the necessity of adopting deep learning-based entity recognition methods.

\subsection{Entity Normalization via Large Language Model}
To enhance entity recognition, we designed a multi-stage entity normalization framework based on the Large Language Model (LLM) \cite{llm-med-natmed} and Retrieval-Augmented Generation (RAG) \cite{llm-rag}. 
First, we packaged each identified symptom or vaccine entity with its surrounding tweet context into a query prompt and used it as input to retrieve relevant standardized concepts. We utilized 260 symptom codes from the 11th revision of the International Classification of Diseases (ICD-11\footnote{ICD: \url{https://icd.who.int/en}}) as the standardization dictionary for symptoms and 47 vaccine generic names as the standardization dictionary for vaccines. 

In the retrieval stage, we employed multiple domain-specific pre-trained language models, including BioLORD \cite{remy-etal-2023-biolord}, S-PubMedBert \cite{deka2022improved}, and BioSimCSE-BioLinkBERT \cite{kanakarajan-etal-2022-biosimcse}, to retrieve 45 coarse-grained candidate concepts for each query. Next, we used the BGE-Gemma-reranker \cite{li2023making} to re-rank the coarse-grained candidate concepts, and selected the top 15 as the fine-grained candidate concepts.

In the generation stage, we merged the query prompt with the 15 fine-grained candidate concepts into a new prompt and input it into the Qwen-1.5-32B-chat model \cite{qwen}, which selected the most appropriate standardized term from the candidate concepts. To prevent error propagation, we introduced a special ``\textit{no normalized noun}'' category in the prompt design. When LLM could not find a suitable standardized term or determined that the entity was not a symptom or vaccine, it would return to this category. After normalization processing, we finally obtained 70,642 symptom entities and 221 corresponding standardized terms, as well as 5,452 vaccine entities and 18 corresponding standardized terms. To evaluate the normalization performance, we randomly sampled 100 entities and their standardized terms for manual inspection, achieving an accuracy of 71\%. 
Table \ref{nn_result} presents the most common physical and psychological symptoms, including their normalized symptoms and raw entities in social media texts. It validates that our method effectively standardizes the informal descriptions found in social media into clinically used ICD codes, facilitating the extraction of useful information for medical research.

\begin{figure*}[hbt]
    \centering
        \includegraphics[width=0.98\textwidth]{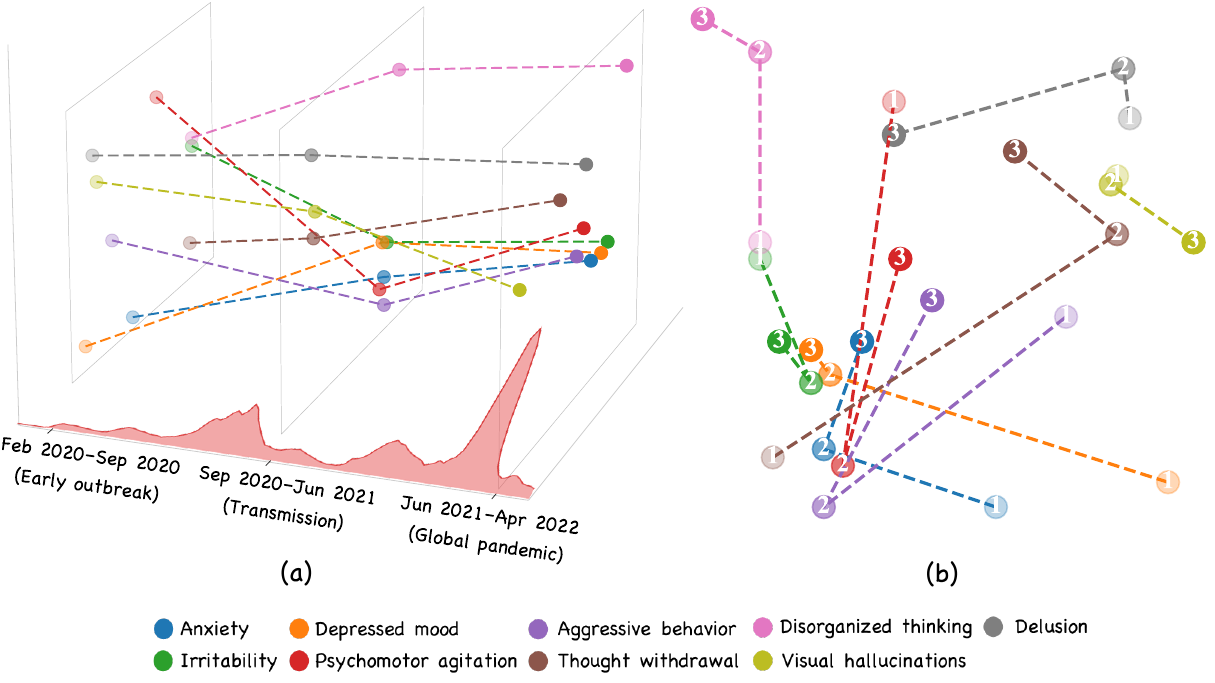}
    \caption{Psychological symptom semantic shift trajectories across different periods. Lines and dots of different colors represent the semantic shift trajectories of various symptoms. Our framework automatically identifies three significant stages of semantic shift, which correspond to the three phases of the pandemic as compared to the number of new COVID-19 cases reported by the WHO.
    (a) shows the semantic relationships of these symptoms at different stages of the pandemic. 
    (b) visually presents the distribution of symptoms in semantic space across different periods, illustrated using gradients of text and color. Each point in the projection graph represents the semantic position of a symptom at a specific period. 
    By analyzing the trajectories, we observe a converging trend as the global spread of the pandemic progresses, indicating increasing semantic association.
    }
    \label{fig:exp}
    \vspace{-0.3cm}
\end{figure*} 


\subsection{Univariate Semantic Shift Analysis}
After obtaining normalized entities for symptoms and vaccines, we further explored their semantic evolution patterns across different stages of the COVID-19 pandemic. 

We divided the tweet data into monthly time slices and extracted word co-occurrence statistics for each time slice using the method proposed in Sections 3.1 and 3.2. Then, following the dynamic update strategy described in Section 3.3, we iteratively updated the word embeddings on each time slice. We obtained three word vector embeddings: Feb 2020-Sep 2020, Sep 2020-June 2021, and June 2021-Apr 2022. 
Based on the number of new COVID-19 cases reported by the WHO\footnote{\url{https://data.who.int/dashboards/covid19/cases}} and the related pandemic timeline\footnote{\url{https://www.cdc.gov/museum/timeline/covid19.html}}, we can roughly correlate these with the three key stages of pandemic development.
The first stage corresponds to the initial outbreak of the epidemic, during which the public gradually begins to pay attention to and discuss the epidemic; the second stage corresponds to the first rapid spread, during which people start to implement control measures; the third stage corresponds to the full-blown outbreak of the epidemic, accompanied by the emergence of two variant viruses. More details are in Appendix \ref{appendix:timeline}. In classic semantic shift problems, anchor words typically align multiple time series into the same embedding space. However, we used Procrustes analysis to align the first two word vectors to the last one, thus obtaining relatively consistent embeddings. 

Procrustes analysis is a linear algebra method used to rotate and translate one matrix onto another while preserving the original structure as much as possible \cite{grave2019unsupervised}. Specifically, given two matrices $A$ and $B$, Procrustes analysis seeks a matrix $R$ such that $A R$ closely approximates $B$ by minimizing the Frobenius norm $\left| AR - B \right|_F$. As it only involves rotation and translation, it does not alter the length or relative positions of word vectors, thus preserving the intrinsic semantic structure of the word vectors. This method ensures that word embeddings from different time slices are aligned in the same semantic space, which helps maintain the global consistency of word vectors and reduces the impact of semantic shift.

We selected psychological symptoms that underwent relatively large changes during the pandemic for univariate semantic shift analysis. As shown in Figure \ref{fig:exp} (a), which introduces temporal dynamics into the univariate semantic shift, and Figure \ref{fig:exp} (b), which projects all semantic shifts onto a plane, we found that with the passage of time, some symptoms such as anxiety, depression, and irritability were more scattered in the early stages of the pandemic and gradually clustered as the pandemic progressed \cite{DepCov-WWW-2023}; some symptoms such as psychomotor agitation and aggressive behavior were initially more clustered and gradually dispersed in the later stages; and some symptoms such as thought withdrawal, thought disorder, delusions, and visual hallucinations remained relatively dispersed throughout \cite{li2022tracking}.

\begin{figure*}[!tp]
    \centering
    \includegraphics[width=\textwidth]{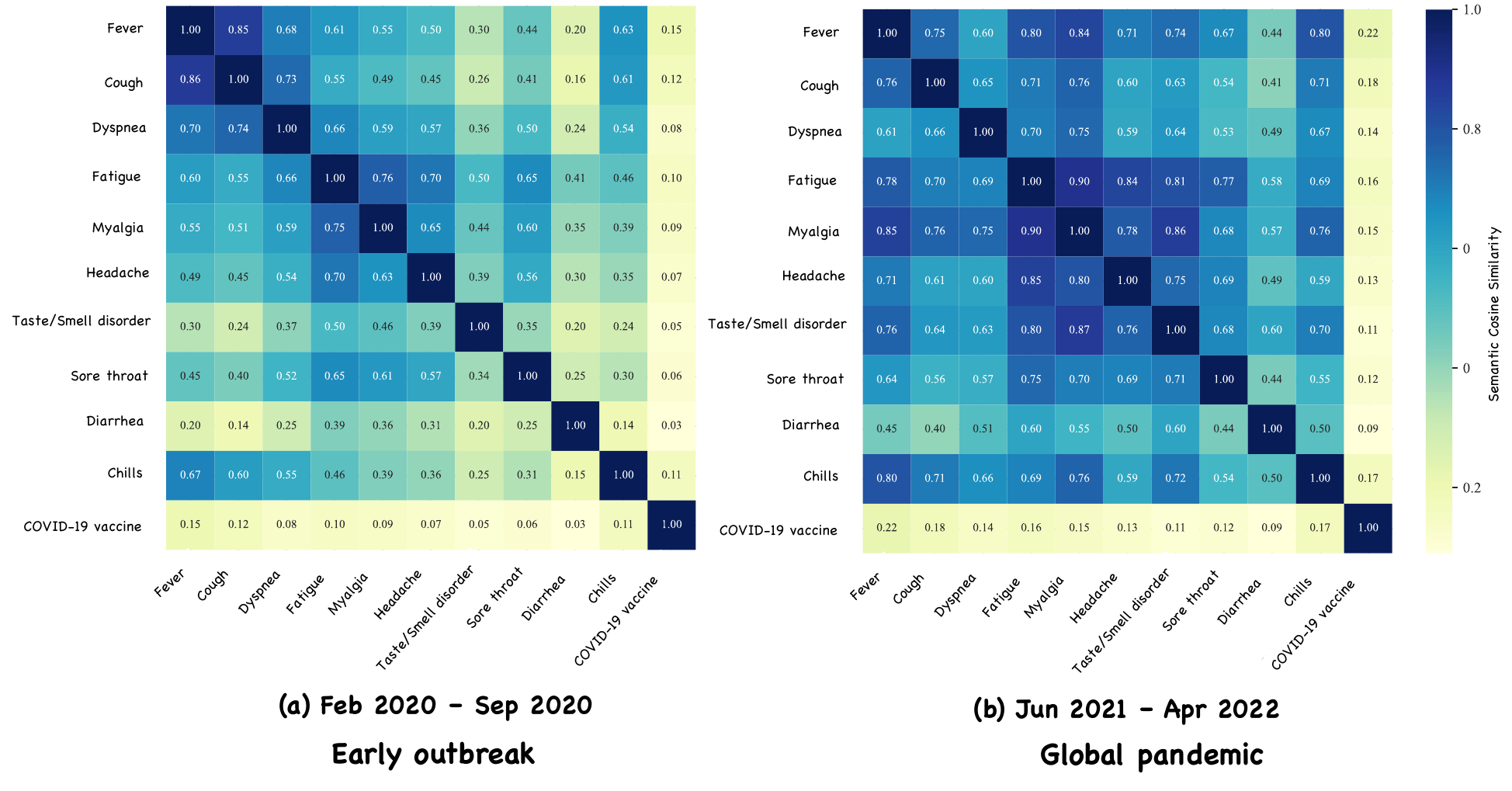}
    \caption{Dynamic longitudinal analysis of COVID-19 symptom-vaccine semantic associations. 
    (a) and (b) show the semantic correlations during the early outbreak (Feb 2020-Sep 2020) and the global pandemic (Jun 2021-Apr 2022), respectively.
    By analyzing the changes in the semantic associations of symptoms-symptoms and symptoms-vaccine across different periods, we examined the potential combination patterns of symptoms and the sensitivity of symptoms to vaccine protection. }
    \label{fig:distance}
    \vspace{-0.3cm}
\end{figure*}

\subsection{Synergistic Effect Analysis} 
To explore the semantic interactions between symptoms and between symptoms and vaccines, we constructed an evolving network of key symptoms and vaccines based on the dynamically updated word embeddings. We used the normalized entities as nodes and calculated their mutual cosine similarity. As shown in Figure \ref{fig:distance}, their inter-relationships were depicted at each time slice.

We analyzed the dynamic evolution patterns and compared them with real-world symptom patterns reported in relevant papers \cite{whitaker2022variant, vihta2023omicron, looi2023covid}. The results showed that the main symptom combinations were closely related to the pandemic stages. The early stage was dominated by respiratory symptoms such as fever and cough, while taste and smell disorders became increasingly prominent in subsequent stages. During the Delta strain period, the proportion of gastrointestinal symptoms such as diarrhea, nausea, and vomiting increased significantly. We further calculated the semantic similarity changes between the vaccine node and various symptom nodes, and found a significant decrease in the association between vaccine administration and multiple key symptoms. This suggests a potential protective effect of the vaccine on these symptoms.

\section{Conclusion and Future work}
This study proposes an innovative unsupervised dynamic word embedding method to capture the longitudinal evolution of word semantics in social media texts. We apply this method to symptom and vaccine semantic analysis using large-scale COVID-19 Twitter data, achieving promising results. Our method can effectively identify the dynamic patterns of symptoms and vaccine semantics at different stages of the epidemic. By introducing temporal dynamics and iterative update strategies, this method overcomes the limitations of traditional static word embeddings, achieving longitudinal tracking of semantic content in social media texts. Additionally, we analyze the semantic interactions between different symptoms and between symptoms and vaccines, revealing their synergistic effects over time.  

In the future, we plan to expand the application of this method in the field of vertical semantic analysis. On one hand, we will explore more advanced time series modeling techniques to more accurately capture the evolving patterns of word semantics over time. On the other hand, we will integrate semantic shift analysis with technologies such as social network analysis and sentiment analysis to investigate the rich semantic relationships and patterns embedded in social media texts, such as identifying precursor signals of sudden events and revealing the driving factors behind the evolution of public opinion.

\section*{Ethics}
Our study uses publicly accessible data collected via Twitter's official API, adhering to the stringent requirements. 
The research utilized tweets obtained following Twitter's Privacy Policy, which informs users that their social media content, including profiles and tweets, is public and freely accessible to third parties
To protect privacy, we removed usernames and only analyzed their tweets. 

\section*{Limitations}
There are several limitations in our work: 1) Assessing the accuracy and effectiveness of semantic shift detection methods is inherently challenging due to the lack of normalized evaluation metrics and annotated datasets. The validation of our method relies on case studies and empirical analysis, which may not generalize well enough across different domains or applications; 2) The initial word embeddings are pre-trained using traditional static word embedding methods. The quality and biases inherent in these initial embeddings could influence the subsequent dynamic updates and potentially the detection of semantic shifts.

\bibliography{custom}

\clearpage
\appendix
\section*{Appendix}

\subsection*{Timeline of COVID-19}
\label{appendix:timeline}
\begin{figure}[ht]
    \centering
    \includegraphics[width=\textwidth]{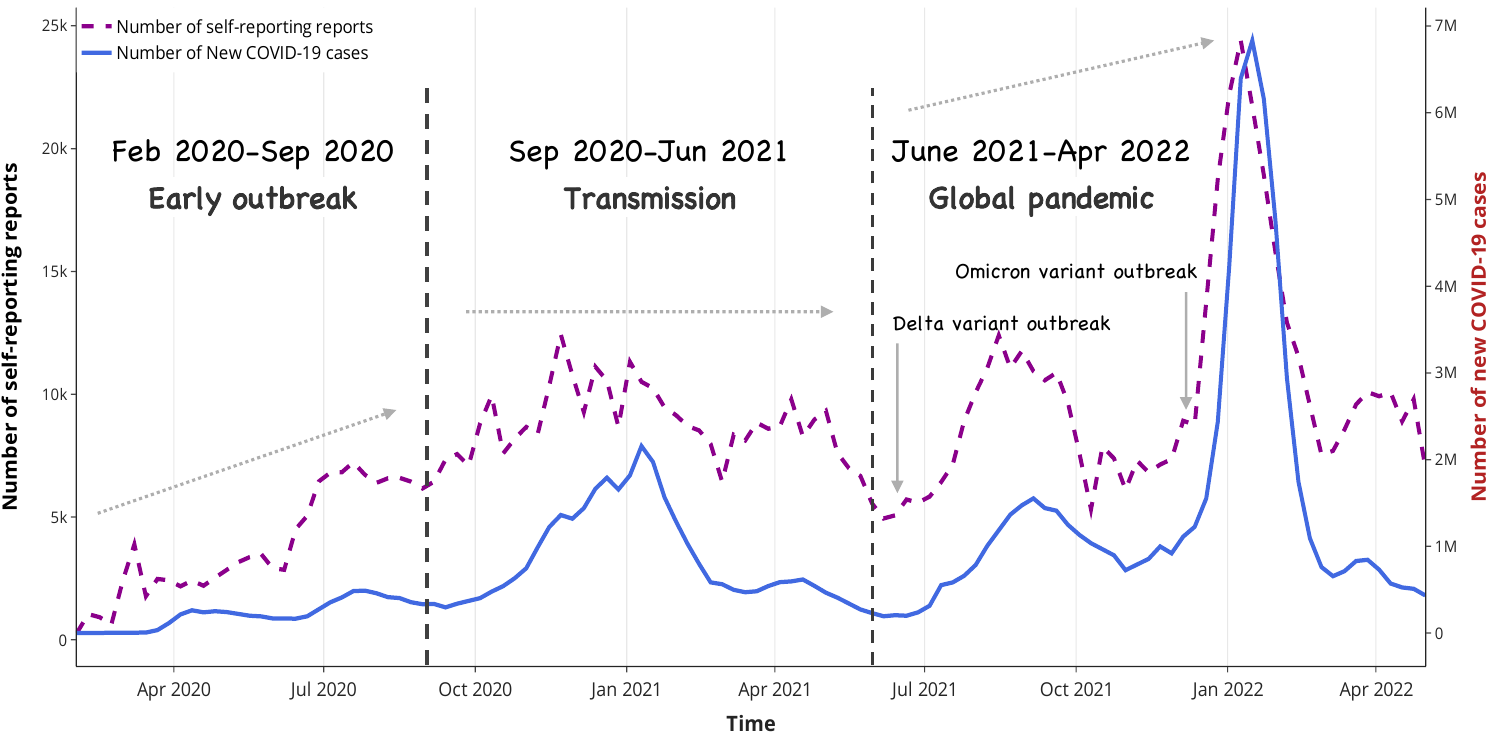}
    \caption{COVID-19 timeline.}
    \label{fig:timeline}
\end{figure}

Figure \ref{fig:timeline} shows the weekly numbers of self-reported COVID-19 tweets and the total new COVID-19 cases in the four main countries of Twitter users (United States, United Kingdom, Canada, and the Philippines) as reported by the WHO.

\end{document}